\documentclass[final,5p,times,twocolumn]{elsarticle}

\usepackage{booktabs} 
\usepackage{multirow} 
\usepackage{colortbl}
\usepackage{xcolor} 
\usepackage{caption}
\usepackage{rotating}

\usepackage{subfig}
\usepackage{lineno,hyperref}
\modulolinenumbers[1]
\usepackage{lineno}

\begin{document}

\begin{frontmatter}

\title{UniMiB SHAR: A Dataset for Human Activity Recognition Using Acceleration Data from Smartphones}

\author{Daniela Micucci, Marco Mobilio and Paolo Napoletano}
\address{Department of Informatics, System and Communication, University of Milano Bicocca, Viale Sarca 336, 20126 Milan, Italy}
\fntext[myfootnote]{Corresponding Author: micucci@disco.unimib.it}




\begin{abstract}
Smartphones, smartwatches, fitness trackers, and ad-hoc wearable devices are being increasingly used to monitor human activities. 
Data acquired by the hosted sensors are usually processed by machine-learning-based algorithms to classify human activities. The success of those algorithms mostly depends on the availability of training (labeled) data that, if made publicly available, would allow researchers to make objective comparisons between techniques. Nowadays, publicly available data sets are few, often contain samples from subjects with too similar characteristics, and very often lack of specific information so that is not possible to select subsets of samples according to specific criteria. 
In this article, we present a new dataset of acceleration samples acquired with an Android smartphone designed for human activity recognition and fall detection. The dataset includes 11,771 samples of both human activities and falls performed by 30 subjects of ages ranging from 18 to 60 years. Samples are divided in 17 fine grained classes grouped in two coarse grained classes: one containing samples of 9 types of activities of daily living (ADL) and the other containing samples of 8 types of falls. The dataset has been stored to include all the information useful to select samples according to different criteria, such as the type of ADL performed, the age, the gender, and so on. 
Finally, the dataset has been benchmarked with four different classifiers and with two different feature vectors. We evaluated four different classification tasks: fall vs no fall, 9 activities, 8 falls, 17 activities and falls. For each classification task we performed a subject-dependent (5-fold cross validation) and a subject-independent (leave-subject-out) evaluation. 
The major findings of the evaluation are the following: i) it is more difficult to distinguish between types of falls than types of activities; ii) subject-dependent evaluation outperforms the subject-independent one. The database will allow researchers to work to more robust features and classification schemes that permit to deal with different types of falls and with effect of data personalization.
\end{abstract}


\end{frontmatter}


\section{Introduction}\label{sec:introduction}

Nowadays, many people lead a \emph{sedentary life} due to the facilities that the increasingly pervasive technologies offer.
Unfortunately, it is recognized that insufficient physical activity is one of the 10 leading risk factors for global mortality: people with poor physical activity is subjected to a risk of all-cause mortality that is 20\% to 30\% higher then people performing at least 150 minutes of moderate intensity physical activity per week~\cite{WHO2010-website}. 
Another important global phenomenon actually affecting our society is \emph{population aging}: the decline or even decrease of the natural population growth rates due to a rise in life expectancy~\cite{UN2013-website} and to a long-term downtrend in fertility (expecially in Europe~\cite{grant_low_2004}). Falls are a major health risk that impacts the quality of life of elderly people. Indeed, among elderly people, accidental falls occur frequently: the 30\% of the over 65 population falls at least once per year; the proportion increases rapidly with age~\cite{tromp2001}. Moreover, fallers who are not able to get up more likely require hospitalization or, even worse, die~\cite{tinetti1993}. 


Thus, research on techniques able to recognize activities of daily living (ADLs), also known as human activities (HA), and to detect falls is very active in recent years: the recognition of ADLs may allow to infer the amount of physical activity that a subject perform daily, while a prompt detection of falls may help in reducing the consequence (even fatal) that a fall may cause mostly in elderly people.

ADLs recognition and fall detection techniques usually accomplish their task by analizing samples from sensors, which can be physically deployed in the ambient (ambient sensors, e.g., cameras, vibration sensors, and microphones) or worn by people (wearable sensors, e.g., accelerometers, gyroscopes, and barometers)~\cite{Mubashir_2013}. To train and evaluate their techniques, researchers usually build their own dataset of samples and rarely make it publicly available~\cite{chen2006,li2009,Ojetola_2015}. This practice makes difficult to compare in an objective way the several newly proposed techniques and implementations due to a lack of a common source of data~\cite{Leutheuser_DALIAC_2013,Ojetola_2015,Vavoulas_MobiFall_2014}. Only very recently, Janidarmian et al. combined 14 publicly available datasets focusing on acceleration patterns in order to conduct an analysis on feature representations and classification techniques for human activity recognition~\cite{Janidarmian_2017}. Unfortunately, they do not make the resulting dataset available for downloading.

The few publicly available datasets can been primary divided into three main sets: acquired by ambient sensors, acquired by wearable devices, and a combination of the two. Recently, a lot of attention has been paid to wearable sensors because they are less intrusive, work outdoors, and often cheaper than the ambient ones. 
This is confirmed by the increasing number of techniques that are based on wearable sensors (see for example the survey by Luque et al. related to fall detection techniques relying on data from smartphones~\cite{Luque2014}). 

Wearable sensors are divided in two main groups: ad-hoc wearable devices (e.g., SHIMMER sensor nodes), and smartphones (e.g., Android). For what concerns fall detection, several studies concluded that, in order to be used, fall detection devices must not stigmatize people nor disturb their daily life~\cite{Steele2009,Feldwieser2016,Jeffs2016}.  Unfortunately, devices such as ad-hoc wearable devices and ambient sensors are not well accepted by elderly people because mostly of their intrusiveness. On the contrary, smartphones are good candidate devices for hosting fall detection systems: they are  are widespread and daily used by a very large number of person, included elderly people. 
This, on the one hand, reduces costs, and on the other, eliminates the problem of having to learn how to use a new device. Moreover, studies demonstrated that samples from smartphones sensors (e.g., accelerometer and gyroscope) are accurate enough to be used in clinical domain, such as ADLs recognition~\cite{Mourcou2015}. This is also confirmed by the amount of publications that rely on the use of smartphone as acquisition devices for fall detection systems~\cite{Vilarinho2015,Luque2014} and ADLs recognition.

For these reasons we concentrate our attention to smartphones as acquisition devices both for ADL recognition and fall detection.
Thus, we searched the publicly available datasets acquired with smartphones in order to identify their strengths and weaknesses so as to outline an effective method for carrying out a new acquisition campaign. We searched the most common repository (IEEE, ACM, Google, and Google Scholar) by using in our query the terms \emph{ADL dataset} and  \emph{Fall dataset} in combination with the following words \emph{smartphone}, \emph{acceleration}, \emph{accelerometer}, \emph{inertial}, \emph{IMU}, \emph{sensor},  and \emph{wearable}. We selected the first 100 results for each query. Removing duplicate entries, we obtained less then 200 different references. Then we manually examined the title, the abstract, and the introduction to eliminate references unrelated to ADL recognition and fall detection, and references that were based on ambient sensors such as camera, microphones, or RFID tags. We then read carefully the remaining references and discarded those that do not make publicly available the dataset used in the experimentation. Finally, we added the relevant references that we missed with our searches but were cited in the papers we selected. At the end of the process, we individuated 13 datasets with data from smartphones and 19 with data from wearable ad-hoc sensors. We then included only those datasets that have been recorded starting from 2012\footnote{We considered the year in which the dataset has been made available. This year do not necessarily coincide with the year in which the related article has been published.} mostly because the oldest dataset including samples from smartphones is dated 2012. This choice  makes the datasets homogeneous with respect to the sensors technologies related to acquisition sensors which rapidly evolves year by year. 
%
%

At the end of the process, we individuated 13 datasets with data from smartphones and 13 with data from wearable ad-hoc sensors. In the following, we will detail some relevant characteristics of the 13 datasets from smartphones since our aim was to build a new dataset containing acceleration patters from smartphone able to complement the existing ones. As it will presented in Section~\ref{sec:dataset}, the datasets from ad-hoc wearable devices have be examined with the aim of identifying the most common ADLs and falls.


Table~\ref{table:datasets} shows the publicly available datasets recorded by means of smartphones and their characteristics. Table~\ref{table:datasets} also includes the dataset we realized in the last row, in order to ease the comparison.

\addtolength{\tabcolsep}{-2pt}  
\begin{table*}[htbp]
\scriptsize
  \centering
    \begin{tabular}{lcccccccccc}
    \toprule
    \multicolumn{1}{l}{\multirow{2}[1]{*}{\textbf{Dataset}}} & 
    \multicolumn{1}{c}{\multirow{2}[1]{*}{\textbf{Year}}} & 
    \multicolumn{1}{c}{\multirow{2}[1]{*}{\textbf{ADLs}}} & 
    \multicolumn{1}{c}{\multirow{2}[1]{*}{\textbf{Falls}}} & 
      \multicolumn{1}{c}{\multirow{2}[1]{*}{\textbf{Nr. of subjects}}} & 
    \multicolumn{2}{c}{\textbf{Gender}} & 
    \multicolumn{1}{c}{\textbf{Age}} &
     \multicolumn{1}{c}{\textbf{Height}} & \textbf{Weight} \\
          &      &       &       &       & \multicolumn{1}{c}{\textbf{Female}} & \multicolumn{1}{c}{\textbf{Male}} & \multicolumn{1}{c}{\textbf{(years)}} & \multicolumn{1}{c}{\textbf{(cm)}} & \textbf{(Kg)} \\
    \toprule
    DMPSBFD~\cite{DMPSBFD_DS} & 2015  &  \cellcolor{lightgray}yes &  \cellcolor{lightgray}yes &  \cellcolor{lightgray}5   & -     & -     & -     & -     & - \\
    \midrule
    
    Gravity~\cite{Vilarinho2015} & 2016  & \cellcolor{lightgray} yes &  \cellcolor{lightgray}  yes &  \cellcolor{lightgray} 2 &   - &  - & \cellcolor{lightgray}  26 - 32 &  \cellcolor{lightgray} 170 - 185 &  \cellcolor{lightgray} 63 - 80 \\
    &       &        &    &   &    &    &\cellcolor{lightgray}  29 $\pm$  4.2    &\cellcolor{lightgray} 178 $\pm$  10.6    & \cellcolor{lightgray}  71.5 $\pm$  12  \\

    \midrule

    MobiFall~\cite{Vavoulas_MobiFall_2014} & 2014  & \cellcolor{lightgray} yes &  \cellcolor{lightgray}  yes &  \cellcolor{lightgray} 24 &  \cellcolor{lightgray}  7 &  \cellcolor{lightgray}  17 & \cellcolor{lightgray}  22 - 47&  \cellcolor{lightgray} 160 - 189 &  \cellcolor{lightgray} 50 - 103 \\
    &       &        &    &   &    &    &\cellcolor{lightgray}  27 $\pm$  5    &\cellcolor{lightgray} 175 $\pm$  7    & \cellcolor{lightgray}  76.4 $\pm$  14.5  \\

    \midrule
   MobiAct~\cite{Vavoulas_MobiAct_2016} & 2016  & \cellcolor{lightgray} yes &  \cellcolor{lightgray}  yes  & \cellcolor{lightgray} 57 &\cellcolor{lightgray} 15 & \cellcolor{lightgray} 42 & \cellcolor{lightgray} 20 - 47 & \cellcolor{lightgray} 160 - 193 & \cellcolor{lightgray} 50 - 120 \\
   &       &        &    &   &    &    &\cellcolor{lightgray}  25 $\pm$  4    &\cellcolor{lightgray} 175 $\pm$  4    & \cellcolor{lightgray}  76.6 $\pm$  14.4 \\

\midrule

    RealWorld (HAR)~\cite{sztyler2016onbody} & 2016  & \cellcolor{lightgray} yes & no &  \cellcolor{lightgray} 16 &  \cellcolor{lightgray}  7 &  \cellcolor{lightgray}  8 & \cellcolor{lightgray}  16 - 62&  \cellcolor{lightgray} 163 - 183 &  \cellcolor{lightgray} 48 - 95 \\
    &       &        &    &   &    &    &\cellcolor{lightgray}  32 $\pm$  12    &\cellcolor{lightgray} 173 $\pm$  7    & \cellcolor{lightgray}  74.1 $\pm$  13.3  \\

   \midrule
    Shoaib PA~\cite{Shoaib_2013} & 2013  &  \cellcolor{lightgray} yes  & no   & 4     &  \cellcolor{lightgray}0 &  \cellcolor{lightgray}4 & 25 - 30 & -     & - \\
    &       &        &    &   &    &    & - &&  \\

  \midrule
   Shoaib SA~\cite{Shoaib_2014} & 2014  &  \cellcolor{lightgray} yes  & no   & \cellcolor{lightgray} 10    &  \cellcolor{lightgray} 0 &   \cellcolor{lightgray} 10 & 25 - 30& -     & - \\
&       &        &    &   &    &    & - &&  \\

    \midrule
    tFall~\cite{Medrano_detecting_2014} & 2013  & yes   &  \cellcolor{lightgray} yes & \cellcolor{lightgray} 10    & 7     & 3     & 20 - 42  & 161 - 184  & 54 - 98  \\
    &       &        &    &   &    &    & 31 $\pm$  9   & 173 $\pm$  1  &  69.2 $\pm$  13.1 \\
    \midrule
    UCI HAR~\cite{Anguita_public_2013} & 2012  &  \cellcolor{lightgray} yes  & no   & \cellcolor{lightgray} 30    & -    & -     & 19 - 48 & -     & - \\
    &       &        &    &   &    &    & - &&  \\

    \midrule
   UCI HAPT~\cite{Reyes_UCI_HAR_updated-version_2016} & 2015  &  \cellcolor{lightgray} yes  & no  & \cellcolor{lightgray}30    & -     & -     & 19 - 48 & -     & - \\
   &       &        &    &   &    &    & - &&  \\

\midrule
    UCI UIWADS~\cite{Casale_UCI_ActivityRec_2012} & 2013  &  \cellcolor{lightgray} yes  & no   & \cellcolor{lightgray} 22    & -    & -     & - & -     & - \\
    &       &        &    &   &    &    & - &&  \\

\midrule
   UMA Fall~\cite{Casilari_UMAFall_2016} & 2016  &  \cellcolor{lightgray} yes  & \cellcolor{lightgray}  yes  &\cellcolor{lightgray} 17 &\cellcolor{lightgray} 6 & \cellcolor{lightgray} 11 & \cellcolor{lightgray} 14 - 55 & \cellcolor{lightgray} 155 - 195 & \cellcolor{lightgray} 50 - 93 \\
   &       &        &    &   &    &    &\cellcolor{lightgray}  27 $\pm$  10    &\cellcolor{lightgray} 172 $\pm$  9    & \cellcolor{lightgray}  69.9 $\pm$  12.3 \\

    \midrule
    WISDM~\cite{Kwapisz_WISDM_2011} & 2012  &\cellcolor{lightgray}yes & no  & \cellcolor{lightgray} 29    & -     & -     & -     & -     & - \\
     \midrule
     
   \textbf{\underline{UniMiB SHAR}} & 2016  & \cellcolor{lightgray} yes &  \cellcolor{lightgray} yes &  \cellcolor{lightgray} 30  &  \cellcolor{lightgray} 24  &  \cellcolor{lightgray} 6  &  \cellcolor{lightgray} 18 - 60  &  \cellcolor{lightgray}160 - 190 &  \cellcolor{lightgray} 50 - 82  \\
    &       &        &    &   &    &    &\cellcolor{lightgray}  27 $\pm$  11    &\cellcolor{lightgray} 169 $\pm$  7     & \cellcolor{lightgray}  64.4 $\pm$  9.7\\
    \bottomrule
    \end{tabular}%
    \caption{The publicly available datasets containing samples from smartphones sensors}
\label{table:datasets}%
\end{table*}%
\addtolength{\tabcolsep}{2pt}

The total number of datasets decreases to 11 because MobiAct and UCI HAPT are updated versions of MobiFall and UCI HAR respectively. Thus, in the following we will refer to 11 datasets overall, discarding MobiFall and UCI HAR. 

The 11 datasets have been recorded in the period 2012 to 2016 (column \emph{Year}). 
Only 5 datasets out of 11 contain both falls (column \emph{Falls}) and ADLs (column \emph{ADLs}). 
%

The average number of subjects for dataset is 18 (column \emph{Nr. of subjects}). The datasets that specify the gender of the subjects (which are MobiAct, RealWorld (HAR), Shoaib PA, Shoaib SA, tFall, and UMA Fall) contain in mean 6 women and 13 men (columns \emph{Gender - Female} and \emph{Gender - Male} respectively). 

DMPSBFD, UCI UIWADS, and WISDM do not specify the age of the subjects (column \emph{Age}). In the remaining 8 datasets, subjects are aged between 21 and 43 on average with a standard deviation of 4 and11respectively. 

Finally, only Gravity, MobiAct, RealWorld (HAR), tFall, and UMA Fall datasets provide detailed information about the height and the weight of the subjects (columns \emph{Height} and \emph{Weight} respectively).

The detailed information reported in Table~\ref{table:datasets} have been collected from the web site hosting the dataset, the readme files of each dataset, and the related papers. It is remarkable to notice that in many cases such information get lost in the downloaded dataset. Grey cells in Table~\ref{table:datasets} indicate that samples are stored so that they can be filtered according to the information contained in the cell. 
%
%
For instance, in all the datasets, with the exception of tFall, it is possible to select subsets of samples according to the specific ADL (column \emph{ADLs}). For example, it is possible to select all the samples that have been labeled \emph{walking}. tFall is an exception because the samples are simply labeled as generic ADL, thus not specifying which specific kind of ADL are. 

For what concerns falls (column \emph{Falls}), all the datasets have organized samples maintaining the information related to the specific type of fall they are related to (e.g., forward).

As specified in column \emph{Nr. of subjects}, the samples are linked to the subjects that performed the related activities and, where provided, falls. This means that in all the datasets (with the exception of Shoaib PA) it is possible to select samples related to a specific subject. However, this information is unhelpful if there is no information on the physical characteristics of the subject. Looking at the double column \emph{Gender}, only MobiAct, RealWorld (HAR), Shoaib PA, Shoaib SA, and UMA Fall maintain information related to the gender of the subject. Finally, it is surprising that only Gravity, MobiAct, RealWorld (HAR), and UMA Fall allow to select samples according to age, height, and/or weight of the subjects (columns \emph{Age}, \emph{Height}, and \emph{Weight}).

In view of this analysis, only MobiAct, RealWorld (HAR), and UMA Fall allow to select samples according to several dimensions, such as the age, the sex, the weight of the subjects, or the type of ADL.  MobiAct and UMA Fall allow to select samples also according to the type of fall. Unfortunately, the other datasets are not suitable in some experimental evaluations. For example, the evaluation of the effects of personalization in classification techniques~\cite{s16010117} taking into account the physical characteristics of the subjects, that is, operating leave-one-subject-out cross-validation~\cite{Khan2017168}.
 


To further contribute to the worldwide collection of accelerometer patterns, in this paper we present a new dataset of smartphone accelerometer samples, named UniMiB SHAR (University of Milano Bicocca Smartphone-based Human Activity Recognition). The dataset was created with the aim of providing the scientific community with a new dataset of acceleration patterns captured by smartphones to be used as a common benchmark for the objective evaluation of both ADLs  recognition and fall detection techniques. 

The dataset has been designed keeping in mind on one side the limitations of the actual publicly available datasets, and on the other the characteristics of MobiAct, RealWorld (HAR), and UMA Fall, so to create a new dataset that juxtaposes and complements MobiAct, RealWorld (HAR), and UMA Fall with respect to the data that is missing. Thus, such a dataset would have to contain a large number of subjects (more than the 18 in average), with a large number of women (to compensate MobiAct, RealWorld (HAR), and UMA Fall), with subjects over the age of 55 (to extend the range of UMA Fall\footnote{We do not consider RealWorld (HAR) since it contains ADLs only. Indeed, it is most difficult to recruit elderly subjects performing falls.}), with different physical characteristics (to maintain heterogeneity), performing a wide number of both ADLs and falls (to be suitable in several contexts). Moreover, the dataset would have to contain all the information required to select subjects or ADLs and falls according to different criteria, such as for example, all the female whose height is in the range 160-168 cm, all the men whose weight is in the range 80-110 Kg, all the walking activities of the subjects whose age is in the range 45-60 years. 


To fulfil those requirements, we built a dataset including 9 different types of ADLs and 8 different types of falls. The dataset contains a total of 11,771 samples describing both activities of daily living (7,579) and falls (4,192) performed by 30 subjects, mostly females (24), of ages ranging from 18 to 60 years.. Each sample is a vector of 151 accelerometer values for each axis.
 Each accelerometer entry in the dataset maintains the information about the subject that generated it. 
Moreover, each accelerometer entry has been labeled by specifying the type of ADL (e.g., \emph{walking},  \emph{sitting}, or  \emph{standing}) or the type of fall (e.g.,  \emph{forward},  \emph{syncope}, or  \emph{backward}).


We benchmarked the dataset by performing several experiments. We evaluated four classifiers: k-Nearest Neighbour (k-NN), Support Vector Machines (SVM), Artificial Neural Networks (ANN), and Random Forest (RF). Raw data and magnitudo have been considered as feature vectors. Finally, for each classification we performed  a subject-dependent (5-fold cross validation) and a subject-independent (leave-subject-out) evaluation. Results show how much the proposed dataset is challenging with respect to a set of classification tasks. 

The article is organized as follows. Section~\ref{sec:dataset} describes the method used to build the datasets. Section~\ref{sec:evaluation} presents the dataset evaluation and Section~\ref{sec:results} discusses the results of the evaluation. Finally, Section~\ref{sec:rconclusion} provides final remarks.

\section{Dataset Description}\label{sec:dataset}
This section describes the method used to acquire and pre-process samples in order to produce the UniMiB SHAR dataset.

\subsection{Data acquisition}


The smartphone used in the experiments was a Samsung Galaxy Nexus I9250 with the Android OS version 5.1.1 and equipped with a Bosh BMA220 acceleration sensor. This sensor is a triaxial \emph{low-g} acceleration sensor. It allows measurements of acceleration in three perpendicular axes, and allows acceleration ranges from $\pm2g$ to $\pm16g$ and sampling rates from 1KHz to 32Hz. The Android OS both limits to $\pm2g$ with a resolution of $~0.004g$ the acceleration range, and takes samples at a maximum frequency of 50Hz. However,  the Android OS does not guarantee any consistency between the requested and the effective sampling rate. Indeed, the acquisition rate usually fluctuates during the acquisition.
For the experiments presented in this paper, we resampled the signal in order to have a constant sampling rate of 50 Hz, which is commonly used in literature for activity recognition from data acquired through smartphones \cite{Medrano_detecting_2014, Shoaib_2014, Anguita_public_2013}. 
The accelerometer signal is for each time instant made of a triplet of numbers (x, y, z) that represents the accelerations along each of the 3 Cartesian axes.

We used also the smartphone built-in microphone to record audio signals with a sample frequency of 8,000 Hz, which are used during the data annotation process.

The subjects were asked to place the smartphone in their front trouser pockets: half of the time in the left one and the remaining time in the right one. 


Acceleration triples and corresponding audio signals have been recorded using a mobile application specially designed and implemented by the authors, which stores data into two separated files inside the memory of the smartphone.

\subsection{ADLs and Falls}

In order to select both the ADLs and the falls, we analyzed the datasets in Table~\ref{table:datasets} and the most recent publicly available datasets recorded with wearable ad-hoc devices. As discussed in Section~\ref{sec:introduction}, we considered the datasets acquired from 2012 to be compliant with the year of the older smartphone-based dataset. This set includes, sorted by year of creation from the oldest the most recent, the following datasets: DLR v2~\cite{Korbinian_DLR_2010}, Ugulino~\cite{Ugulino2012}, USC HAD~\cite{mi12:ubicomp-sagaware}, DaLiAc~\cite{Leutheuser_DALIAC_2013}, EvAAL~\cite{Kozina_2013}, MHEALTH~\cite{Banos_MHEALTH}, UCI ARSA~\cite{Casale_UCI_ActivityRec_2012}, BaSA~\cite{Leutheuser_BaSA_2014}, UR Fall Detection~\cite{KWOLEK2014489}, MMsys~\cite{Ojetola_2015}, SisFall~\cite{Sucerquia_SisFall_2017}, UMA Fall\footnote{UMA Fall contains samples from both smartphones and ad-hoc wearable devices.}~\cite{Vilarinho2015}, and REALDISP~\cite{Banos_REALDISP}.

\begin{figure*}[htbp]
\begin{center}
\includegraphics[width=14cm]{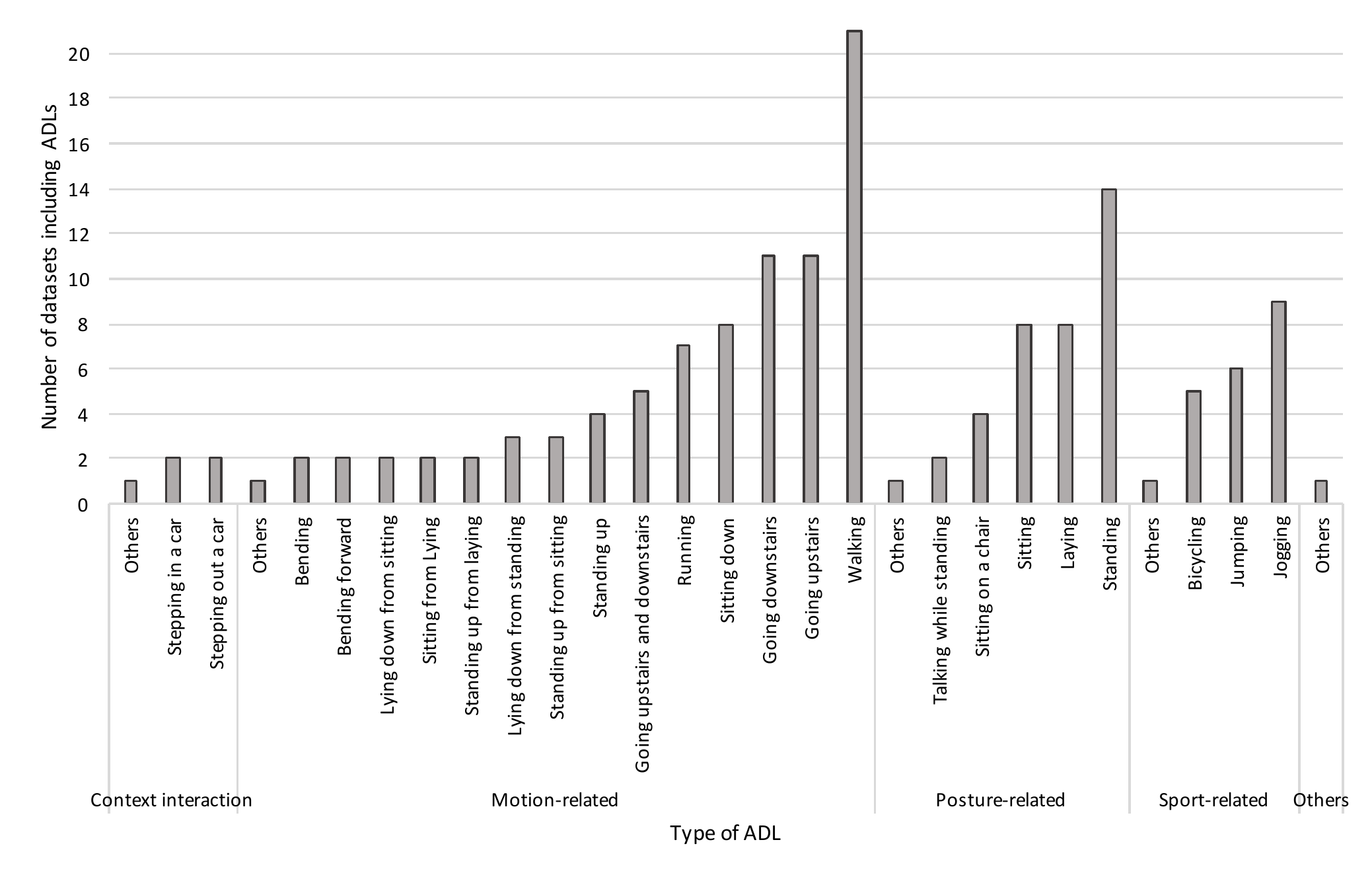}
\caption{ADLs and theirs occurrence in the publicly available datasets analysed grouped by category}.
\label{fig:ADLs}
\end{center}
\end{figure*}

For what concerns ADLs, Figure~\ref{fig:ADLs} shows the most common ones in the overall 24 datasets we analyzed (11 with samples from smartphones and sketched in Table~\ref{table:datasets} and 13 with samples from wearable ad-hoc devices listed above). The y axis represents the number of datasets that include the specified ADL. ADLs are grouped by category. The following categories have been identified by analizyng the datasets: \emph{Context-related}, which includes activities that someway deal with the context (e.g., \emph{Stepping in a car}), \emph{Motion-related}, which includes activities that imply some kind of physical movement (e.g., \emph{Walking}), \emph{Posture-related}, which includes activities in which the person maintains the position for a certain amount of time (e.g., \emph{Standing}), \emph{Sport-related}, which includes any kind of activity that requires a physical effort  (e.g., \emph{Jumping}), and \emph{Others}, which includes activities that are presented in one dataset only (e.g., \emph{Vacuuming} in category \emph{Housekeeping-related}). The \emph{Jogging} and \emph{Running} activities deserve a clarification. In all the datasets we analyzed, they are mutually exclusive, that is, datasets that contain \emph{Running}, do not contain \emph{Jogging} and vice versa. The datasets REALDISP and MHEALT are an exception because they include both the activities. These datasets, besides being realized by the same institution, are primarily oriented towards the recognition of physical activities (warm up, cool down and fitness exercises). Moreover, none of the datasets analyzed exactly specify what the \emph{Jogging} and \emph{Running} activities are related to. Thus, even though they may be considered very similar activities, we have decided to keep them separated in oder to do not loose their specificity. We classify \emph{Jogging} as a \emph{sport-related} activity (in the sense, for instance, of jogging in the park), and \emph{Running} as a \emph{motion-related} activity (in the sense, for instance, of running for the bus).
For each category, the x axis shows all the ADLs we found and that are present in at least 2 datasets. Under the label \emph{Others} fall all the ADLs for the corresponding category that have been included in one dataset only (e.g., \emph{Walking left-circle} in category \emph{Motion-related}).
%
%
%

Tables~\ref{table:adls} shows the 9 ADLs we have selected among the most popular included in the analyzed publicly available datasets. UniMiB SHAR includes the top 5 most popular \emph{Motion-related} activities (i.e., \emph{Walking}, \emph{Going upstairs}, \emph{Going downstairs}, \emph{Sitting down}, and \emph{Running}). 
Moreover, we detailed the generic \emph{Standing up}, by including the \emph{Standing up from sitting} and \emph{Standing up from laying} activities. Finally, we included also the \emph{Lying down from standing}.

In the \emph{Sport-related} category, we did not included \emph{Jogging}  even if it is the most popular activity in its category because we included the \emph{Running} activity in the \emph{Motion-related} category. 
In \emph{Sport-related} category, we chose the \emph{Jumping} activity being the second most popular one.

Our dataset does not include \emph{Postural-related} activities. Indeed, we were interested in acquiring acceleration data from activities related to movements both because from them it is possible to estimate the overall physical activity performed by a person, and because people are more likely to fall during movements~\cite{Robinovitch2013}.

We do not include ADLs belonging to categories such as \emph{Housekeeping-},  \emph{Cooking-}, or \emph{Personal care-related} (those fall in the \emph{Others} category in Table~\ref{fig:ADLs}, because we are interested in \emph{\bf{low order}} activities of daily living, which include simple activities such as, \emph{Standing}, \emph{Sitting down}, \emph{Walking}, rather than \emph{\bf{high order}} activities of daily living, which include complex activities such as, \emph{Washing dishes}, \emph{Combing hair}, \emph{Preparing a sandwich}. The same holds for \emph{contex-realted} activities that are intended as  \emph{\bf{high order}} activities. This choice was also motivated by the fact that these activities are scarcely present in the analyzed datasets (in particular, each activity belonging in the above mentioned categories is present in only one of the 24 analyzed datasets).

Finally, among the ADLs related to movements, we selected the ADLs most used in literature as demonstrated by the analysis we performed, which is also confirmed by Pannurat et al. in \cite{Natthapon2014}.

\begin{table*}[tb]
  \centering
\resizebox{0.8\textwidth}{!}{
    \begin{tabular}{llll}
    \textbf{Category} & \textbf{Name} & \textbf{Description} & \textbf{Label}\\
    \toprule
    Motion-related & Standing up from laying & From laying on the bed to standing&StandingUpFL\\
    
     & Lying down from standing & From standing to lying on a bed &LyingDownFS\\
    
    & Standing up from sitting & From standing to sitting on a chair &StandingUpFS\\
     & Running & Moderate running &Running\\
    & Sitting down & From standing to sitting on a chair &SittingDown\\
    & Going downstairs &  Climb the stairs moderately&GoingDownS\\
    & Going upstairs & Down the stairs moderately&GoingUpS \\
    & Walking & Normal walking &Walking\\
    \midrule
    Sport-related & Jumping & Continuos jumping &Jumping\\
    \bottomrule
    \end{tabular}%
    }
  \caption{ADLs performed by the subjects In the UniMiB SHAR dataset}
  \label{table:adls}%
\end{table*}%

For what concerns falls, we analized DMPSBFD, Gravity, MobiAct, tFall, and UMAFall datasets from smartphones (see Table~\ref{table:datasets}), and DLR v2, EvAAL, MMsys, SISFall, UMA Fall, and UR Fall Detection datasets from wearable ad-hoc devices, since they are the only datasets that contain falls. From this set, we excluded  DLR v2, EvAAL because they do not specify the type of fall.

\begin{figure*}[tb]
\begin{center}
\includegraphics[width=14cm]{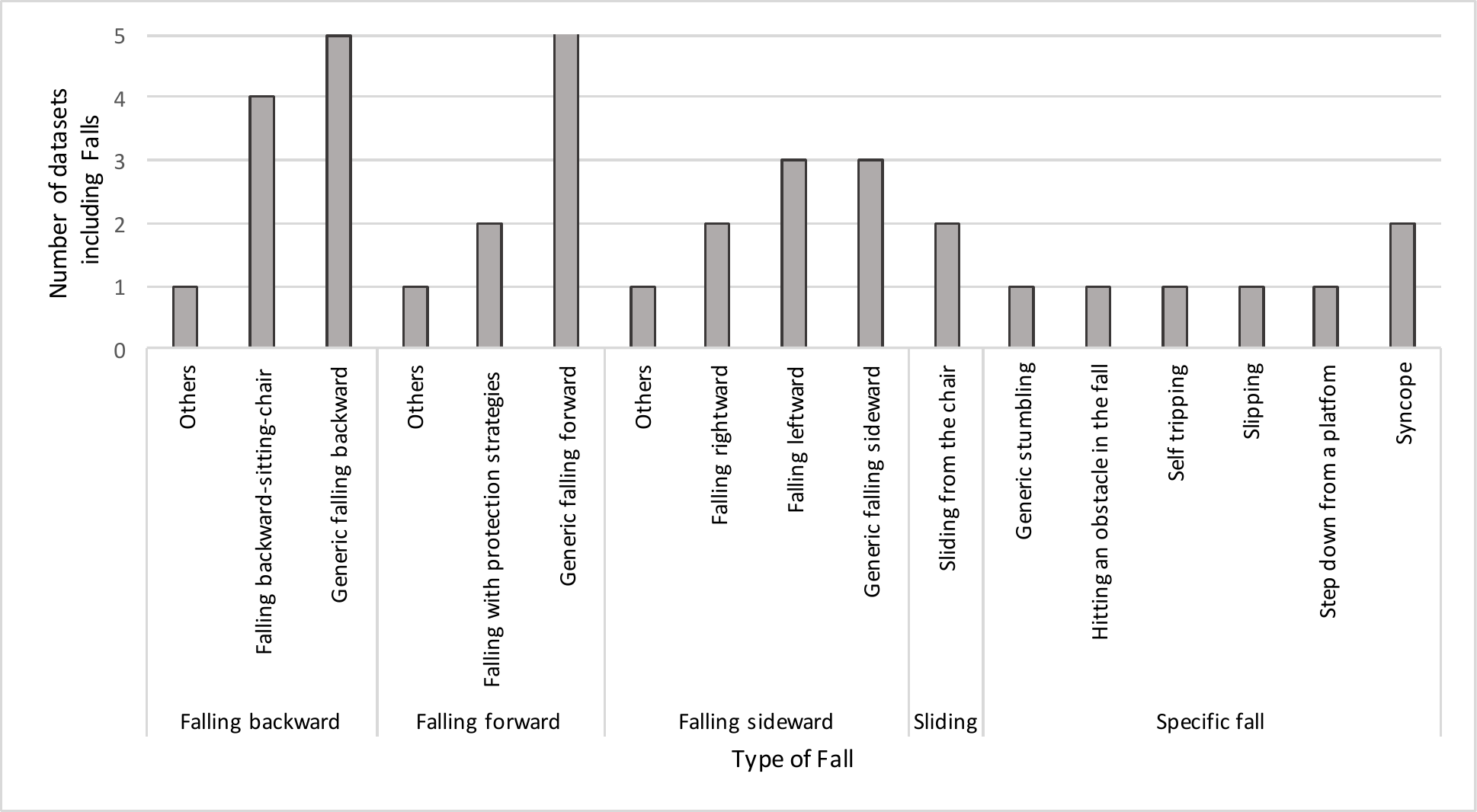}
\caption{Falls and theirs occurrence in the publicly available datasets analysed grouped by category}.
\label{fig:Falls}
\end{center}
\end{figure*}

Figure~\ref{fig:Falls} shows the most common falls in the resulting 9 datasets we analyzed. The y axis represents the number of datasets that include the specified fall. Likewise ADLs, falls are grouped by category. \emph{Falling backward},  \emph{Falling forward}, and \emph{Falling sideward} include back-, front-, and side-ward falls respectively. \emph{Sliding} category can be further specialized so that to include \emph{Sliding from a chair}, \emph{Sliding form a bed}, and \emph{Generic sliding} that not specifies details about the type of sliding. Finally, the category \emph{Specific fall} includes different type of falls that have not been further specialized. 

For each category, the x axis shows all the types of falls we found. Under the label \emph{Others} fall all the falls for the corresponding category that have been included in one dataset only. The \emph{Specific fall} category is an exception since it includes falls types not particularly present in the analyzed datasets.

\begin{table*}[htbp]
\resizebox{\textwidth}{!}{%
  \centering
    \begin{tabular}{llll}
    \textbf{Category} & \textbf{Name} & \textbf{Description} &\textbf{Label}\\
    \toprule
    Falling backward  & Falling backward-sitting-chair  & Fall backward while trying to sit on a chair &FallingBackSC\\
    & \emph{Generic} falling backward & Generic fall backward from standing &FallingBack\\
     
    \midrule
    Falling forward & Falling with protection strategies  & Falls using compensation strategies to prevent the impact& FallingWithPS\\    
     & \emph{Generic} falling forward & Fall forward from standing, use of hands to dampen fall &FallingForw\\

    \midrule
    Falling sideward & Falling rightward & Fall right from standing &FallingLeft \\
     & Falling leftward & Fall right from standing &FallingRight\\
    \midrule
    
    Specific fall & Hitting an obstacle in the fall & Falls with contact to an obstacle before hitting the ground &HittingObstacle \\
    & Syncope & Getting unconscious&Syncope \\
      \bottomrule
    \end{tabular}
 }
   \caption{Falls performed by the subjects in teh UniMiB SHAR dataset}\label{table:falls}
\end{table*}%

Choosing which falls to include in the dataset was driven by the following considerations: the number of falls should have been comparable to that of the other datasets, and the dataset should have included a set of representative types of falls. Thus, having four categories (not considering \emph{Sliding}, which includes only one type of fall that has been considered by two datasets only), we selected two falls from each of them. In each category, we selected the first two most popular falls. The category \emph{Falling sideward} is an exception since we preferred to choose the two most specific falls instead of including the too generic \emph{Generic falling sideward}. 
%
Table~\ref{table:falls} shows the 8 falls that we selected according the adopted criterion.

Finally, studies on this topic confirm the falls we selected are common in real-life~\cite{Igual_comparison_2015,Kangas2012500,Klenk2011368,Medrano_detecting_2014}. 

\subsection{Subjects}
30 healthy subjects have been involved in the experiments: 24 were women and 6 men. The subjects, whose data are shown in Table~\ref{table:subjects}, are aged between 18 and 60 years (27 $\pm$12 years), have a body mass between 50 and 82 kg (64.4 $\pm$ 9.7 kg), and a height between 160 and 190 cm (169 $\pm$ 7 cm). Note that we included more women and older ages to compensate for the lacks of MobiAct.

\begin{table}[htbp]
  \centering
    \scriptsize
  
    \begin{tabular}{lcccc}
          & \textbf{Total} & \textbf{Female} & \textbf{Male} &  \\
\cmidrule{1-4}    \textbf{subjects} & 30    & 24    & 6     &  \\
    \midrule
    \multirow{2}[4]{*}{\textbf{age}} & 18 - 60 & 18 - 55 & 20 - 60 & min - max \\
\cmidrule{2-5}          & 27 $\pm$ 11 & 24 $\pm$ 9 & 36 $\pm$ 15 & mean $\pm$ std \\
    \midrule
    \multirow{2}[4]{*}{\textbf{height}} & 160 -190 & 160 - 172 & 170 - 190 & min - max \\
\cmidrule{2-5}          & 169 $\pm$ 7 & 166 $\pm$ 4 & 179 $\pm$ 6 & mean $\pm$ std \\
    \midrule
    \multirow{2}[4]{*}{\textbf{weight}} & 50 - 82 & 50 - 78 & 55 - 82 & min - max \\
\cmidrule{2-5}          & 64.4 $\pm$ 9.7 & 61.9 $\pm$ 7.8 & 74.7 $\pm$ 9.7 & mean $\pm$ std \\
    \bottomrule
    \end{tabular}%
 \caption{The characteristics of the subjects}\label{table:subjects}%
\end{table}%

All the subjects performed both ADLs and Falls. The subjects gave written informed consent and the study was conducted in accordance with the WMA Declaration of Helsinki~\cite{WMA2013-helsinki}.


\subsection{Protocols}


To simplify the data annotation process, we asked each subject to clap her hands early before and after she performed the activity/fall to be recorded. 
Moreover, to reduce background noise, we asked each subject to wear gym trousers with front pockets. 

Concerning ADLs, in order to avoid mistakes by the subjects  due to too long sequences of activities, 
registrations have been subdivided in the three protocols showed in Table~\ref{table:ADL-protocol}. Each protocol has been performed by each subject twice, the first one with the smartphone in the right pocket and the second in the left.  
Those smartphone positions were chosen because both they are the most natural ones and they are exactly the positions used in the analyzed references dealing with smartphones. 

\begin{table}[htbp]
  \centering
  \scriptsize
    \begin{tabular}{lll}
    \textbf{Protocol} & \textbf{Action} & \textbf{Iteration} \\
    \midrule
    \multirow{9}[2]{*}{Protocol 1} & Start the registration & \multirow{9}[2]{*}{1 time} \\
          & Put the smartphone in the pocket &  \\
          & \textit{\textbf{clap}} &  \\
          & Walking for 30 seconds &  \\
          & \textit{\textbf{clap}} &  \\
          & Running for 30 seconds &  \\
          & \textit{\textbf{clap}} &  \\
          & Pull the smartphone from the pocket &  \\
          & Stop the registration &  \\
    \midrule
    \multirow{13}[2]{*}{Protocol 2} & Start the registration & \multirow{13}[2]{*}{1 time} \\
          & Put the smartphone in the pocket &  \\
          & \textit{\textbf{clap}} &  \\
          & Climb 15 steps &  \\
          & \textit{\textbf{clap}} &  \\
          & Go down 15 steps &  \\
          & \textit{\textbf{clap}} &  \\
          & Wait 2 seconds &  \\
          & \textit{\textbf{clap}} &  \\
          & Jump 5 times &  \\
          & \textit{\textbf{clap}} &  \\
          & Pull the smartphone from the pocket &  \\
          & Stop the registration &  \\
    \midrule
    \multirow{16}[8]{*}{Protocol 3} & Start the registration & \multirow{2}[2]{*}{1 time} \\
          & Put the smartphone in the pocket &  \\
\cmidrule{2-3}          & \textit{\textbf{clap}} & \multirow{6}[2]{*}{5 times} \\
          & Sitting down &  \\
          & \textit{\textbf{clap}} &  \\
          & Standing up from sitting &  \\
          & \textit{\textbf{clap}} &  \\
          & \textit{Wait 2 seconds} &  \\
\cmidrule{2-3}          & \textit{\textbf{clap}} & \multirow{6}[2]{*}{5 times} \\
          & Lying down from standing &  \\
          & \textit{\textbf{clap}} &  \\
          & Standing up from laying  &  \\
          & \textit{\textbf{clap}} &  \\
          & \textit{Wait 2 seconds} &  \\
\cmidrule{2-3}          & Pull the smartphone from the pocket & \multirow{2}[2]{*}{1 time} \\
          & Stop the registration &  \\
    \bottomrule
    \end{tabular}%
    \caption{The protocols for ADLs aquisition}
\label{table:ADL-protocol}%
\end{table}%

\emph{Protocol 1} includes \emph{Walking} and \emph{Running} activities. We opted for moderate walking and running so as to include even older people. \emph{Protocol 2} includes activities related to both climbing and descending stairs, and jumps. In our registration, we selected straight stairs ramps, and asked each volunteer to perform jumps with a moderate elevation, with little effort, and spaced each other about 2 seconds. \emph{Protocol 3} includes ascending and descending activities. The \emph{Sitting down} and \emph{Standing up from sitting} activities have been performed with a chair without armrests; the \emph{Lying down from standing} and \emph{Standing up from laying} have been performed on a sofa. The duration of the actives are in average with those reported in~\cite{s17071513} that reviews a set of public datasets for wearable fall detection systems.

Falls have been recorded individually, always following the pattern of making a start and end clap (see Table~\ref{table:fall-protocol}). In cases where the volunteer ended in a prone position, the clap has been performed by an external subject to avoid as far as possible any movements that might lead to recording events outside the study. To carry out the simulation safely, a mattress of about 15 centimeters in height was used. Each fall was repeated six times, the first three with the smartphone in the right pocket, the others in the left. Finally, falls have been simulated, started from a standing straight up position, and self-started.
\begin{table}[htbp]
  \centering
    \begin{tabular}{ll}
    \textbf{Action} & \textbf{Iteration} \\
    \midrule
    Start the registration & \multirow{7}[2]{*}{6 times} \\
    Put the smartphone in the pocket &  \\
    \textit{\textbf{clap}} &  \\
    fall  &  \\
    \textit{\textbf{clap}} &  \\
    Pull the smartphone from the pocket &  \\
    Stop the registration &  \\
    \bottomrule
    \end{tabular}%
    \caption{Protocol for each fall}\label{table:fall-protocol}%
\end{table}%

\subsection{Segmentation and preprocessing}
The audio files helped in the identification of the start and stop time instants for each recorded activity.
From the labelled recorded accelerometer data, we extracted a signal window of 3 \emph{sec} each time a peak was found, that is, when the following conditions were verified:
\begin{enumerate} 
\item the magnitude of the signal $m_t$ at time $t$ was higher than $1.5g$, with $g$ being the gravitational acceleration; 
\item the magnitude $m_{t-1}$ at the previous time instant $t-1$ was lower than $0$.
\end{enumerate} 

 Each signal window of 3 \emph{sec} was centered around each peak and it is likely that several overlap between subsequent windows may happen. We adopted this segmentation technique instead of selecting overlapped sliding windows because our dataset is mostly focused on motion-related recognition of ADLs and falls.  The choice of taking 3 \emph{sec} window has been motivated by: i) the cadence of an average person walking is within [90, 130] steps/min~\cite{anguita2013public,benabdelkader2002stride}; ii) at least a full walking cycle (two steps) is preferred on each window sample. 

\begin{figure*}[tb]
\begin{center}
\includegraphics[width=14cm]{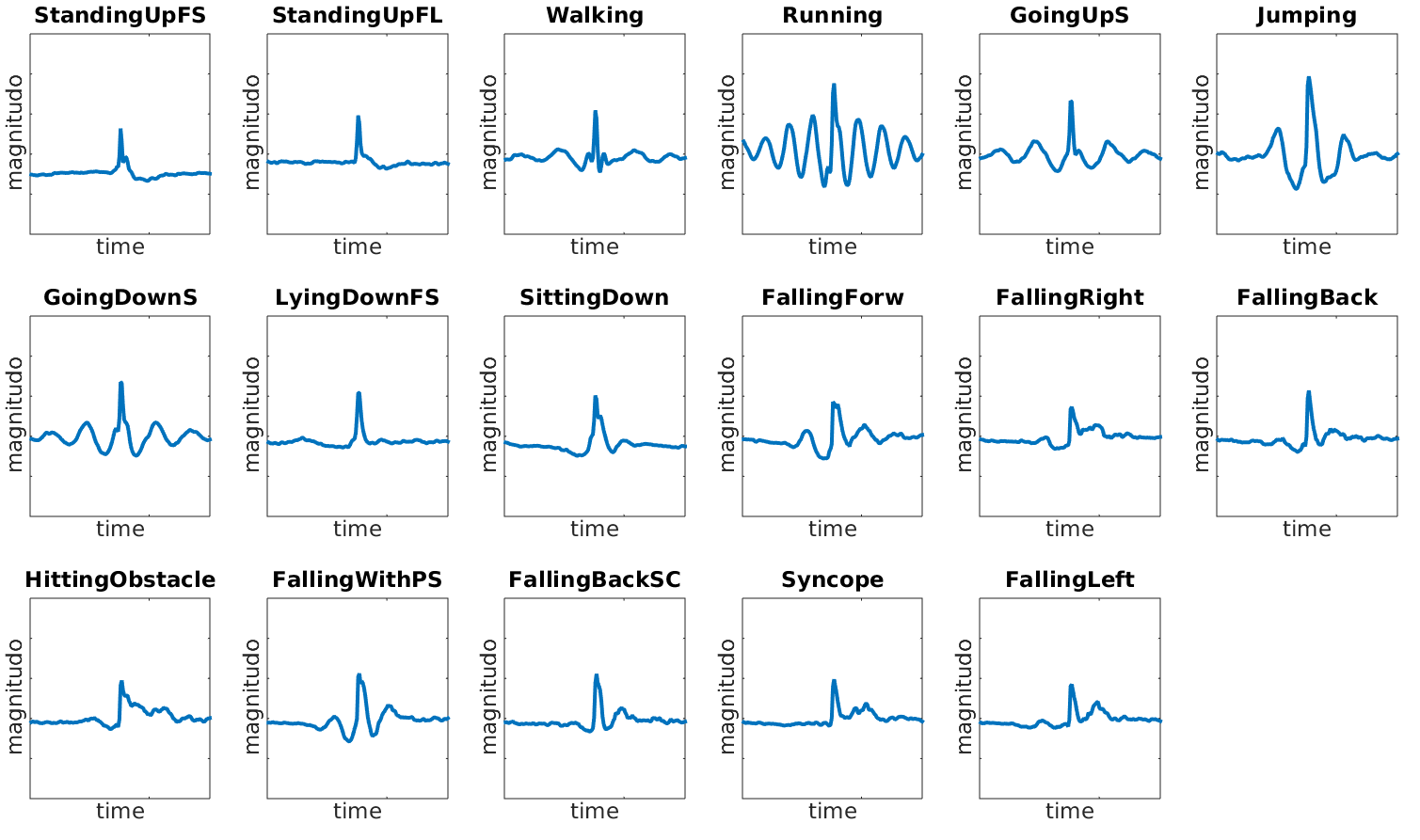}
\caption{Samples of acceleration shapes}
\label{fig:activitysample}
\end{center}
\end{figure*}

Figure~\ref{fig:activitysample} shows samples of acceleration shapes. For each activity, we displayed the average magnitude shape obtained by averaging all the subjects' shapes.

\begin{figure}[tb]
\begin{center}
\includegraphics[width=9cm]{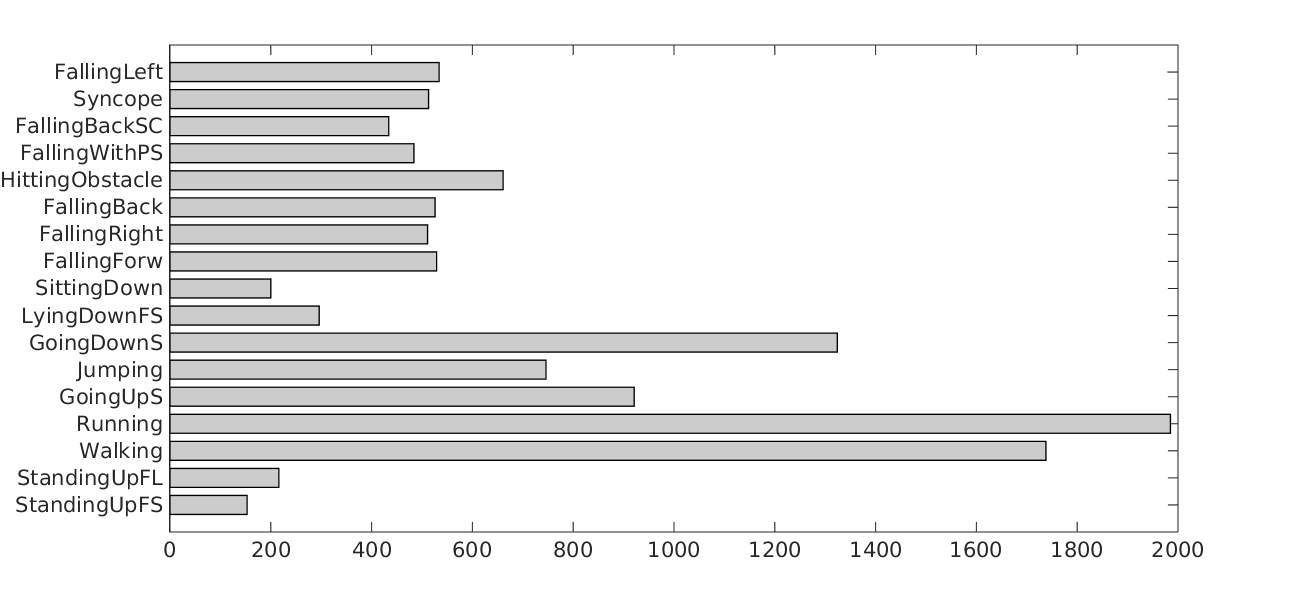}
\caption{Activity samples distribution}
\label{fig:distribution}
\end{center}
\end{figure}

Since the device used for data acquisition records accelerometer data with a sample frequency of 50 Hz, for each activity, the accelerometer data vector is made of 3 vectors of 151 values (a vector of size 1x453), one for each acceleration direction. The dataset is thus composed of 11,771 samples describing both ADLs (7,759) and falls (4,192) not equally distributed across activity types. This is because the activity of running and walking were performed by subjects for a time longer than the time spent for other activities. Originally, 6,000 time windows of the running activity were found. In order to make the final dataset as much as balanced, we have deleted about 4,000 samples related to running activities.  The resulting samples distribution is plotted in Figure~\ref{fig:distribution}, where the samples related to running activities are about 2,000. On our web site we release both datasets, the one balanced and the original one.

We preprocessed the acceleration signal $s(t)$ in order to remove the gravitational component $g(t)$. Since the gravitational force is assumed to have only low frequency components, we applied a Butterworth ($BW$) low-pass filter with a cut off frequency of 0.3 Hz~\cite{anguita2013public}: $g(t) = BW(s(t),0.3)$. The accelerometer data without gravitational component is then obtained as: $\tilde{s(t)} = s(t)-g(t)$.

\section{Dataset Evaluation}\label{sec:evaluation}

We organized the accelerometer samples in order to evaluate four classification tasks:

\begin{enumerate}
\item  \emph{AF-17} contains 17 classes obtained by grouping all the 9 classes of ADLs and 8 classes of FALLs. This subset permits to evaluate the capability of the classifier to distinguish among different types of ADLs and FALLs;
\item  \emph{AF-2} contains 2 classes obtained by considering all the ADLs as one class and all the FALLs as one class. This subset permits to evaluate, whatever is the type of ADL or FALL, the classifier robustness in distinguishing between ADLs and FALLs;
\item \emph{A-9} contains 9 classes obtained by considering all the 9 classes of ADLs. This subset permits to evaluate how much the classifier is capable to distinguish among different types of ADLs;
\item \emph{F-8} contains 8 classes obtained by considering all the 8 classes of FALLs. This subset permits to evaluate how much the classifier is capable to distinguish among different types of FALLs.
\end{enumerate}

We initially evaluated the classifiers by performing a traditional \emph{5-fold} cross-validation. It means that all the data have been randomly split in 5 folds. Each fold has been considered as test data and the remaining ones as training data. Results are computed by averaging the result obtained on each test fold.  The folds have been obtained by applying the stratified random sampling that ensures samples of the same subject in both the test and the training folds.

To make the dataset evaluation independent from the effect of personalization, we conducted another evaluation by performing a \emph{leave-subject-out} cross-validation. Each test fold is made of accelerometer samples of one user only, namely the \emph{test user}, while the training folds contain accelerometer samples of all the other users except the samples of the \emph{test user}. 

Previous studies demonstrated that classifiers trained on raw data perform better with respect to classifiers trained on other types of feature vector representations, such as magnitude of the signal, frequency, or energy~\cite{Igual_comparison_2015,Micucci2017}. However, in order to make the experiments comparable with others experiments presented in the state of the art, we considered two feature vectors:

\begin{enumerate}
\item \emph{raw data}: the 453-dimensional patterns obtained by concatenating the 151 acceleration values recorded along each Cartesian direction;
\item \emph{magnitude} of the accelerometer signal, that is a feature vector of 151 values.
\end{enumerate}

We experimented  four different classifiers:

\begin{enumerate}
\item k-Nearest Neighbour (k-NN) with $k=1$;
\item Support Vector Machines (SVM) with a radial basis kernel;
\item Artificial Neural Networks (ANN). We set up a three-layers feed forward network with back propagation. The network architecture includes an input layer, a layer of hidden neurons and an output layer that includes a softmax function for class prediction. The number of hidden neurons $n$ has been set in way that $n=\sqrt{m \times k}$, where $m$ is the number of neurons in the input layer and $k$ is the number of neurons in the output layer, namely the number of classes~\cite{xu2016wearable}.
\item Random Forest (RF): bootstrap-aggregated decision trees with 300 bagged classification trees.
\end{enumerate}
All the classifiers have been implemented exploiting the MATLAB Statistics and Machine Learning Toolbox and the Neural Network Toolbox.


\subsection{Evaluation metrics}

As shown in Figure~\ref{fig:distribution}, each of the 17 sets containing samples related to a specific activity is different in size.  To cope with the class imbalance problem of the dataset we used as metric the \emph{macro average accuracy}~\cite{he2013imbalanced}.

Given $E$ the set of all the activities types, $a \in E$, $NP_{a}$ the number of times $a$ occurs in the dataset, 
and $TP_{a}$ the number of times the activity $a$ is recognized, MAA (\emph{Macro Average Accuracy}) is defined by Equation~\ref{eq:MAA}. 
%


\begin{equation}
MAA = \frac{1}{|E|}\sum_{a=1}^{|E|} Acc_{a} = \frac{1}{|E|}\sum_{a=1}^{|E|} \frac{TP_{a}}{NP_{a}}.
\label{eq:MAA}
\end{equation}

MAA is the arithmetic average of the accuracy $Acc_{a}$ of each activity. It  
 allows each partial accuracy to contribute equally to the evaluation.


\section{Results and Discussion}\label{sec:results}

In the following, we discuss separately the results achieved with the traditional 5-fold cross-validation and the leave-subject-out cross-validation.
\subsection{Subject-dependent evaluation (5-fold evaluation)}
The k-fold evaluation is the most employed evaluation scheme in literature~\cite{kittler1982pattern}. This evaluation considers a training set and a test set made of activity samples performed by all the human subjects. The resulting classifier is subject-dependent and usually exhibits a very high performance. Results of the k-fold evaluation (here we used k=5) scheme are showed in Table~\ref{tab:kfold_exp_5fold}  for raw data and magnitude. Overall, the performances achieved using raw data are better than the ones obtained using magnitude as feature vector. This confirm a result already achieved in previous works~\cite{Igual_comparison_2015,Micucci2017}.

\begin{figure}[tb]
\begin{center}
\includegraphics[width=9cm]{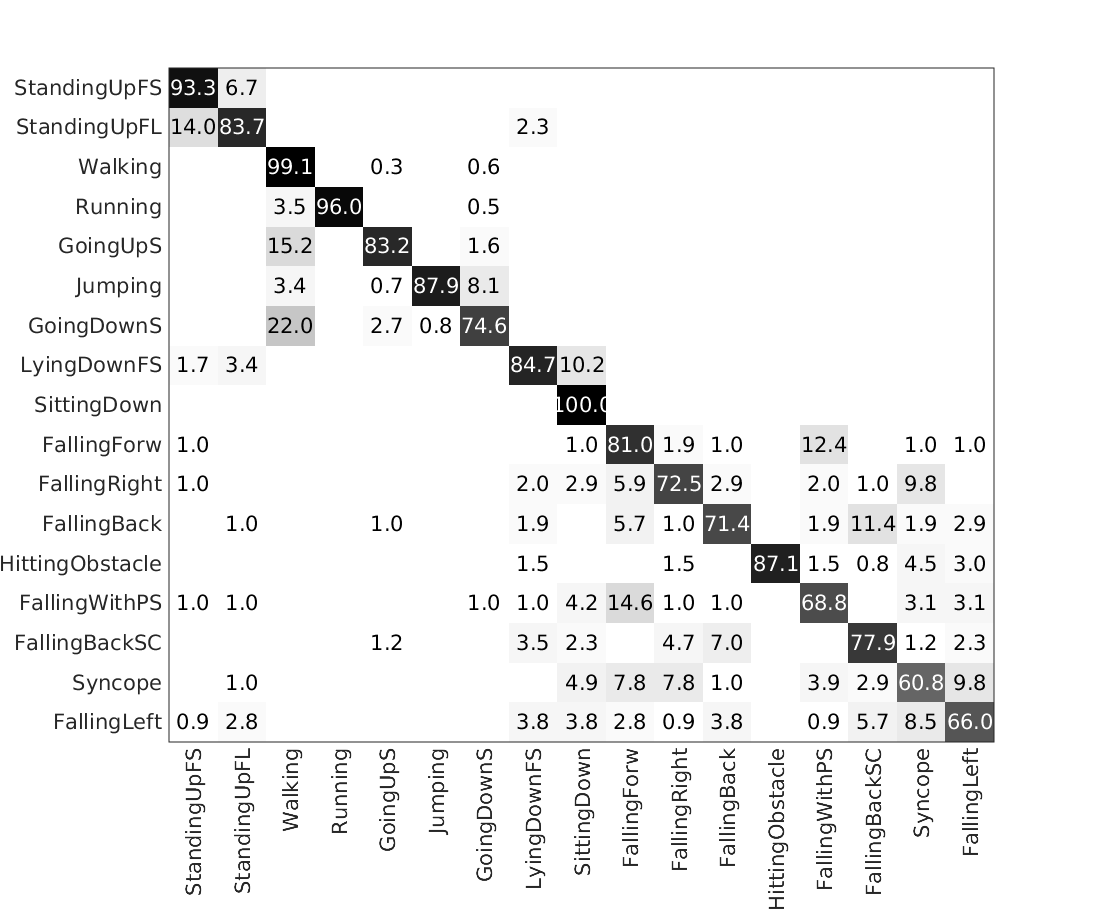}\\
\caption{Confusion matrix of the AF-17 classification achieved with k-NN}
\label{fig:confmat1}
\end{center}
\end{figure}

The AF-17 recognition task is quite challenging with a MAA of about 83\% in the case of raw data with KNN, and a MAA of about 66\% in the case of magnitude with RF. This means that is quite difficult to distinguish among types of activities especially in the case when magnitude is adopted as feature vector.  Figure~\ref{fig:confmat1} shows the confusion matrix of the k-NN experiment in the case of raw data.

The A-9 classification task is quite easy, the MAA obtained by raw data with RF is about 88\% while the MAA obtained by magnitude with SVM is about 79\%. This means that is quite easy to distinguish between types of activities. Looking at the confusion matrix in Figure~\ref{fig:confmat1}, the most misclassified pairs of activities are \emph{Standing up from laying} and  \emph{Standing up from sitting},  \emph{Lying down from standing} and  \emph{Sitting down}, \emph{Going upstairs} and \emph{Walking}, \emph{Going downstairs} and \emph{Walking}, \emph{Jumping} and \emph{Going downstairs}.
 
The F-8 recognition task is quite challenging: the MAA is about 78\% and 57\% in the case of raw data with KNN and magnitude with RF respectively. This result suggests that distinguish among falls is very complicated The most misclassified pairs of falls are \emph{Falling with protection strategies} and  \emph{Generic falling forward}, \emph{Syncope} and \emph{Falling leftward}, \emph{Generic falling backward} and \emph{Falling backward-sitting-chair}, \emph{Falling rightward} and \emph{Falling with protection strategies}, \emph{Falling rightward}  and \emph{Syncope}.

In contrast, the AF-2 recognition task is very easy for all the classifiers and for both raw data and magnitude with a MAA of about 99\% achieved with raw data and SVM. These results are similar to those obtained by previous researchers on a similar classification task performed on different datasets~\cite{Micucci2017,Medrano_detecting_2014}. This means that it is very easy to distinguish between falls and no falls.

To summarize, the F-8 and AF-17 are quite challenging classification tasks. The use of this dataset for those tasks will permit researchers:
\begin{itemize}
\item to design and evaluate more robust feature representations as well as more robust classification schemes for human activity recognition.
\item to study more robust features to deal with accelerometer samples of different types of falls.
\end{itemize}





\begin{table}[tb]
\centering
\scriptsize
{\begin{tabular}{lcccc|cccc}
\multicolumn{9}{c}{5-fold}\\
          \hline
&\multicolumn{4}{c}{\emph{raw data}}&\multicolumn{4}{c}{\emph{magnitude}}\\
          \hline
     data &KNN&SVM&ANN&RF&KNN&SVM&ANN&RF\\
     \hline
AF-17 &\bf{82.86} &78.75 &56.06 &81.48 & 65.30 &65.71 &41.95 & \bf{65.96}\\
AF-2 &97.78 &\bf{98.71} &98.57 &98.09 &95.56 &\bf{97.42} &96.71 &95.74 \\
A-9 &87.77 &81.62 &72.13 &\bf{88.41} &77.37 &\bf{78.94} &62.81 &75.14 \\
F-8 &\bf{78.55} &75.63 &55.07 &78.27 &53.31 &56.34 &37.66 &\bf{57.26} \\
   \hline
\end{tabular}
\caption{5-fold evaluation. Mean Average Accuracy for each classification task using raw data and magnitude as feature vectors. In bold the best results for each classification task and feature vector employed.}\label{tab:kfold_exp_5fold}}{}
\end{table}

\begin{table}[tb]
\centering
\scriptsize
{\begin{tabular}{lcccc|cccc}
\multicolumn{9}{c}{leave-subject-out}\\
          \hline
&\multicolumn{4}{c}{\emph{raw data}}&\multicolumn{4}{c}{\emph{magnitude}}\\
          \hline
     data &KNN&SVM&ANN&RF&KNN&SVM&ANN&RF\\
     \hline
%
 AF-17 &52.14 &55.15 &48.00 &\bf{56.53}& 52.14 &55.09 &48.00 &\bf{56.58} \\
AF-2  &92.90 &\bf{97.57} &95.41&97.02 & 92.90 &\bf{97.57} &96.07 &97.05 \\
A-9  &63.79 &63.32 &63.63 &\bf{73.17}& 63.79 &63.36 &63.63 &\bf{72.67} \\
F-8 &43.66 &\bf{48.84} &38.50 &45.88& 43.66 &\bf{49.35} &38.50 &45.26 \\
     \hline
\end{tabular}
\caption{Leave-subject-out. Mean Average Accuracy for each classification task using raw data and magnitude of the signal as feature vectors. In bold the best results for each classification task and feature vector employed.}\label{tab:kfold_exp_lso}}{}
\end{table}

\subsection{Subject-independent evaluation (leave-subject-out evaluation)}
Table~\ref{tab:kfold_exp_lso} shows the results obtained by performing the leave-subject-out evaluation. In this case the training set is made of activity samples of subjects not included in the test set. This evaluation is also known as subject independent evaluation and shows the feasibility of a real smartphone application for human activity recognition~\cite{medrano2016effect,sztyler2017online,weiss2012impact} where data of a given subject are usually not included in the training set of the classifier.


From the results it is almost evident the drop of performances with respect to the case of 5-fold evaluation. Human subject performs activities in a different way and this influences the recognition accuracy especially when it is necessary to distinguish between fine grained types of activities, that is in the case of AF-17, A-9 and F8 recognition tasks. In particular, in the case of AF-17 the best MAA is 56.58\% using RF and magnitude. In the case of A-9 the best MAA is 73.17\% using RF and raw data. In the case of F-8 the best MAA is 49.35\% using SVM and magnitude. In contrast, distinguishing between coarse grained activities, such as falls vs no falls, is quite easy with a MAA of 97.57\% with SVM for both raw data and magnitude. 
Overall the magnitude feature vector performs slightly better than the case of raw data. This suggests that using the magnitude as feature vector in the case of subject-independent evaluations could be more reliable than raw data.

The low performance achieved in the case of subject-independent evaluation permits researcher to investigate the following issues:
\begin{itemize}
\item the study of a more robust feature vector that is able to reduce as much as possible the performance gap between the subject-dependent and subject-independent evaluation;
\item the study of on-line learning classification schemes that permit, with the use of a few subject-dependent data, to improve as much as possible the performance. 
\end{itemize}

\section{Conclusion}\label{sec:rconclusion}

Almost all publicly available datasets from smartphones do not allow the selection of samples based on specific criteria related to the physical characteristics of subjects and the activities (and/or falls) they performed. Of the 11 datasets containing smartphone measurements, only MobiAct, RealWorld (HAR), and UMA Fall are the exception. These three datasets include more men than women. Considering only datasets that include falls (MobiAct and UMA Fall), the maximum age of the subjects is 47 years. Our goal was therefore to create a new dastaset that would be complementary to the more complete ones and that also include falls. The result is UniMiB SHAR dataset that includes 9 ADLs and 8 falls performed by 30 humans, mostly female, with a huge range of ages, from 18 to 60 years.

The classification results obtained on the proposed dataset showed that the raw data performs quite better than magnitude as feature vector in the case of subject-dependent evaluation, and, on the opposite, the magnitude performs quite better than raw data in the case of subject-independent evaluation. The classification of different types of activities is simpler than the classification of different types of falls. It is very easy to distinguish between falls and no falls for both raw data and magnitude. The subject-independent evaluation showed that recognition performance strongly depends on the subject data.

UniMiB SHAR dataset will permit researchers to study several issues, such as: i) robust features to deal with falls; ii) robust features and classification schemes to deal with personalization issues.

We are planning to carry out an evaluation of the state-of-the-art techniques for ADLs recognition on both UniMiB SHAR and all the publicly available datasets of accelerometer data from smartphone to have and objective comparison. Moreover, we have planned to make experimentation on personalization by using those datasets that include information about the characteristics of the subjects. We want to investigate whether the training set containing samples acquired by subjects with similar characteristics to the testing subject may result in a more effective classifier.  Finally, we are planning to check if and how data from smartwatches and smartphones can jointly improve the performances of the classifiers. To this end, we are improving the data acquisition application used for UniMiB SHAR.

\vspace{6pt} 

\section*{Supplementary}

The dataset, the Matlab scripts to repeat the experiments, the app used to acquire samples, and additional materials (e.g., images with samples of acceleration shapes) are available at the following address: \url{http://www.sal.disco.unimib.it/technologies/unimib-shar/.}
\section*{References}

\bibliography{biblio}


\end{document}